\documentclass[11pt]{article}
\usepackage{acl}
\usepackage{fontspec}
\newfontfamily\cyrfont{LiberationSerif}[Path=fonts/, Extension=.ttf, UprightFont=*-Regular, ItalicFont=*-Italic, BoldFont=*-Bold, BoldItalicFont=*-BoldItalic]
\usepackage{microtype}
\usepackage{booktabs}
\usepackage{graphicx}
\usepackage{amsmath}
\usepackage{xcolor}
\newcommand{\kk}[1]{{\cyrfont\itshape #1}}
\input{results}
\input{results_extra}
\newcommand{\releaseat}{}

\title{Unchecked or Falsely Accused? Evaluating AI-Generated Text Detectors on Kazakh}

\ifdefined\ARRBUILD
\author{Anonymous}
\newcommand{\annotator}{A native-speaker annotator (an author)}
\else
\author{Mansur Ozaman \\
  School of Computing and Artificial Intelligence \\
  Nazarbayev University, Astana, Kazakhstan \\
  \texttt{mansur.ozaman@nu.edu.kz}}
\renewcommand{\releaseat}{\footnote{Code: \url{https://github.com/ozxmn/aina-kk}; data: \url{https://huggingface.co/datasets/ajnasiou/aina-kk}; detector: \url{https://huggingface.co/ajnasiou/aina-kk-det}.}}

\newcommand{\annotator}{The author, a native speaker,}
\fi

\begin{document}
\maketitle

\begin{abstract}
AI-text detectors are already applied to Kazakh writing, yet no study reports how often publicly available detectors flag Kazakh text whose human authorship is dated before ChatGPT. We present AINA-KK: \RESnHuman{} human passages from encyclopedic, news, web and educational pages that provably existed before ChatGPT, and \RESnMachine{} machine and gray-zone passages from \RESnGen{} generators, including the Kazakh-built KazLLM and Sherkala, with script and length stress tests. Five English-trained detectors from the Hugging Face Hub flag up to \res{ots/maxfpr}\% of human passages and \res{ots/max-essay-fpr}\% of published student essays; recalibrated, they catch at most \res{ots/max-tpr-dev}\% of machine text. Calibrated zero-shot detectors catch at most \res{zs/best-tpr}\% of machine text from in-distribution generators. Fine-tuned encoders rank text almost perfectly, but mDeBERTa at its default threshold flags \res{det/base/short/50/fpr-at-0.5}\% of \res{const/trunc-short}-word human excerpts, partly because machine outputs are shorter than human text; calibrated, it catches only \res{det/base/flag-by-gen-at-dev/kazllm}\% of KazLLM and \res{det/base/flag-by-gen-at-dev/sherkala}\% of Sherkala passages. Without script handling, machine-transliterated Latin-2021 text defeats every detector. Fixes that need no new data (calibration, script canonicalization, script-augmented and length-matched training, and an OR with a zero-shot scorer) form AINA-KK-Det. On test it flags \res{x/dep/combo/fpr}\% of human passages (95\% CI \res{x/dep/combo/fpr-ci/lo}--\res{x/dep/combo/fpr-ci/hi}; \res{x/devdep/combo/fpr}\% on development data) and catches \res{x/dep/combo/tpr}\% of machine text from known generators, \res{x/dep/combo/gemma4}\% from \res{corpus/n-heldout-gemma} unseen Gemini-family Gemma-4 models and \res{x/dep/combo/kz}\% from the Kazakh-built ones; its configuration was chosen with test results in view. We release the data, code and detector\releaseat.

\end{abstract}

\section{Introduction}
Detectors of machine-generated text (MGT) are now part of academic-integrity workflows, but almost all of their evidence comes from English and a few high-resource languages. Kazakh, the state language of Kazakhstan, sits in an uncomfortable middle. Turnitin's AI writing detection supports \res{ext/turnitin-n-langs} languages (\res{ext/turnitin-langs}), not Kazakh, and returns no report for other languages.\footnote{\url{https://guides.turnitin.com/hc/en-us/articles/28477544839821} (accessed \res{ext/turnitin-accessed}).} GPTZero, Copyleaks, Originality.ai and Pangram do not list Kazakh among their supported languages (support pages checked \res{ext/turnitin-accessed}). At the same time, a vendor used by Kazakh universities markets Kazakh AI detection with false positives ``reduced to \res{ext/antiplagiat-claim}\%''.\footnote{\url{https://antiplagiat.com/news/system-shows-the-percentage-of-ai-text/} (\res{ext/antiplagiat-date}).} Meanwhile ChatGPT and Gemini account for about \res{ext/statcounter-share}\% of AI-chatbot traffic in the country.\footnote{StatCounter, \res{ext/statcounter-month}: \url{https://gs.statcounter.com/ai-chatbot-market-share/all/kazakhstan}.}

Two failures follow. Kazakh text can pass \emph{unchecked}, or it can be judged by an untested tool that \emph{falsely accuses} honest writers. Kazakh also adds two conditions that English benchmarks never test. It is written in Cyrillic today but is officially moving to a Latin-based alphabet, and real Kazakh text routinely mixes in Latin look-alike letters. And Kazakh now has its own instruction-tuned LLMs \citep{koto2025sherkala,kazllm2024}, whose output may not resemble that of the global models detectors are trained on. Prior work has touched Kazakh only lightly. The GenAI Content Detection shared task at COLING 2025 included \res{ext/coling25-kk-samples} Kazakh test samples, a language unseen in its training data \citep{wang2025genai}. A human study measured \res{ext/human-study-acc}\% accuracy on \res{ext/human-study-pairs} GPT-4o/Wikipedia pairs \citep{wang2026is}. Two small classifier datasets exist, one pairing human reviews with outputs of the Kazakh-built Sherkala \citep{anekesh2026kazaidetect}, the other built from Qwen2.5 outputs and their paraphrases \citep{hardkazakh2026}. None of this work evaluates the public detectors an institution can download on text whose human authorship is dated, tests script, or compares Kazakh-built with global generators.

We contribute:
\textbf{(1) AINA-KK}, a benchmark of \RESnHuman{} human passages whose existence before ChatGPT is provable, in \res{corpus/n-domains} domains including student reports and essays, and \RESnMachine{} machine and gray-zone passages from \RESnGen{} generators, among them the Kazakh-built Sherkala and KazLLM, with script and length stress tests (\S\ref{sec:data}).
\textbf{(2) An evaluation} of zero-shot detectors over \res{corpus/n-zs-families} scoring-model families, fine-tuned multilingual encoders, and widely downloaded public detectors, with false-positive rates by domain and length (\S\ref{sec:findings}). Both failures occur: English-trained public detectors flag up to \res{ots/maxfpr}\% of human writing, calibrated zero-shot detectors catch at most \res{zs/best-tpr}\% of in-distribution machine text, fine-tuned detectors over-flag short human texts, two Kazakh-adapted Llama models evade fine-tuned detectors, and without script handling machine-transliterated Latin text defeats every calibrated threshold.
\textbf{(3) Fixes} that need no new data collection (\S\ref{sec:fixes}): thresholds calibrated on unseen human text and by length, script canonicalization before detection, script-augmented and length-matched training, and an OR-combination of a fine-tuned classifier with a zero-shot scorer. Together they form \textbf{AINA-KK-Det}, a detector we release with the data and code. Adding Kazakh-built generators to training, by contrast, does not reliably fix their evasion.

\section{Related Work}
\paragraph{Multilingual MGT benchmarks.} M4 \citep{wang2024m}, M4GT-Bench \citep{wang2024mgtbench}, SemEval-2024 Task 8 \citep{wang2024semeval}, MULTITuDE \citep{macko2023multitud}, MultiSocial \citep{macko2025multisoc}, CEAID \citep{macko2026ceaid} and DetectRL-X \citep{wu2026detectrl} cover up to \res{ext/max-languages} languages. None includes Kazakh in training; among multilingual benchmarks, Kazakh appears only in the COLING 2025 test set \citep{wang2025genai} and as human fact-check text in BLUFF \citep{lucas2026bluff}. RAID \citep{dugan2024raid} and MAGE \citep{li2024mage} stress-test detectors with adversarial edits and domain shift, in English.
\paragraph{Evasion and gray zones.} Paraphrasing \citep{krishna2023paraphra}, expert edits \citep{artemova2025beemo} and homoglyphs \citep{creo2025silversp,macko2024authorship} evade detection, and detectors flag lightly AI-polished writing \citep{saha2025almost}. Much low-resource web text is itself machine-translated \citep{thompson2024a}, which threatens any ``human'' web corpus.
\paragraph{Detectors and false accusation.} Zero-shot detectors score text with a language model: likelihood and rank \citep{gehrmann2019gltr,su2023detectllm}, DetectGPT \citep{mitchell2023detectgp} and its fast variant \citep{bao2024fastdete}, and Binoculars \citep{hans2024spotting}; supervised detectors train a classifier \citep{verma2024ghostbus,emi2024technica}. Detectors penalize non-native English writers \citep{liang2023gpt,stowe2026identifying}, though a Czech replication found no systematic bias \citep{ali2026differen}, and commercial tools were found unreliable \citep{weberwulff2023testing}. Detection with explicit false-positive guarantees has been proposed \citep{zhu2025reliably,zhou2026detectin}. Short texts are a known weak point \citep{chakraborty2024position}: multiscale training improves detection of short machine text \citep{tian2024multiscale}. We show the converse failure, short \emph{human} text flagged because of a length imbalance in the training data.
\paragraph{Kazakh NLP.} KazMMLU \citep{togmanov2025kazmmlu}, Sherkala \citep{koto2025sherkala}, KazLLM \citep{kazllm2024}, Qor\'{g}au \citep{goloburda2025qorgau}, KazParC \citep{yeshpanov2024kazparc} and TUMLU \citep{isbarov2025tumlu} supply models and benchmarks, but none addresses AI-text detection.

\section{AINA-KK}
\label{sec:data}
\paragraph{Human text with provable provenance.} Every human passage existed before ChatGPT's release (\res{const/cutoff-date}). \emph{Encyclopedic}: the \res{ext/wiki-dump-date} Kazakh Wikipedia dump (archive.org item \texttt{kkwiki-20220501}); much of Kazakh Wikipedia was imported in \res{ext/wiki-import-year} from the state Kazakh Encyclopedia. \emph{News}: pages from \RESnOutlets{} Kazakh-language outlets whose Common Crawl capture predates \res{const/cutoff-date}, from FineWeb2 \citep{fineweb2}. \emph{Web}: CC100-kk \citep{conneau2020xlmr,wenzek2020ccnet}, built from Common Crawl snapshots of \res{ext/cc100-years}. \emph{Educational}: student reports and essays (\kk{реферат}, \kk{шығарма}, \kk{эссе}) on \RESnEduHosts{} educational websites, from the same pre-cutoff FineWeb2 crawl; lesson plans and tests are excluded. Passages are paragraph-aligned spans of \res{const/span-min}--\res{const/span-max} words, one per source page (after the cleaning below, \res{len/h-under120}\% fall under \res{const/span-min} words). We keep passages that GlotLID \citep{kargaran2023glotlid} labels \texttt{kaz\_Cyrl} with confidence $\geq$\res{const/glotlid-p}, remove near-duplicates (MinHash, Jaccard $\geq$\res{const/jaccard}), and scrub emails, phone numbers and \res{const/id-digits}-digit ID numbers. A second pass, derived from the validity audit below, drops leftover wiki markup, breadcrumbs, related-link lists and download banners. Passages are clustered by topic (character TF-IDF, $k$-means) and whole clusters are assigned to train or test, so no topic cluster appears in both; no human test passage has a near-duplicate among training passages even at Jaccard $\geq$\res{const/xsplit-jaccard}.

\paragraph{Machine and gray-zone text.} Prompts are written in Kazakh, with separate template pools for train and test (Table~\ref{tab:gens}). \textbf{C1} topic$\to$text uses the source passage's title or opening clause and the register of its domain (encyclopedia article, news story, school essay or report). \textbf{C2} continues the first $\sim$\res{const/c2-prefix} words of a human passage and also yields mixed-authorship passages (MIX) with a known boundary. \textbf{C3} AI-polishes a human passage (HP). \textbf{C4} (LLM) and \textbf{C5} (NLLB-200; \citealp{nllb2022}) translate pre-2022 Russian web text into Kazakh (MT). \textbf{C6} rewrites machine text with a different model under a ``humanizing'' prompt (MH). C1 and C2 ask for about \res{const/req-words} words, drawn at random. Outputs pass the same language filter plus refusal, repetition and echoed-instruction filters; those under \res{const/min-words} words are dropped and longer ones trimmed to at most \res{const/max-words} words at a sentence end, and we drop outputs in which a single word mixes Cyrillic with another script; this removed \res{qc/mixed-script-drops} outputs, \res{qc/mixed-script-drops-g35fl} of them from Gemini 3.5 Flash-Lite, whose Kazakh output intermittently switches script mid-word. Outputs of \res{corpus/heldout-names} (\res{corpus/n-heldout-rows} passages: \res{corpus/n-heldout-m} machine, \res{corpus/n-heldout-mix} mixed) are held out: no classifier is trained on them except in the last three rows of Table~\ref{tab:fixes}, and we score their outputs for test-split sources (for development sources in Table~\ref{tab:devrep}). Gemma models are built from the technology behind Gemini, two of our training generators, so transfer to them stays within a family; KazLLM and Sherkala, both derived from Llama-3.1-8B, are the only held-out generators from a family absent from training.

\begin{table}[t]\centering\footnotesize\setlength{\tabcolsep}{2.2pt}
\begin{tabular}{llrrrrr}\toprule
Generator & Access & M & MIX & HP & MT & MH\\\midrule
Gemini 3.5 Flash-Lite & Gemini & 831 & 388 & 0 & 0 & 147\\
Gemini 3.1 Flash-Lite & Gemini & 758 & 351 & 127 & 86 & 0\\
Gemma 4 31B$^\dagger$ & Gemini & 404 & 185 & 0 & 0 & 0\\
gpt-oss-120b & Groq & 316 & 149 & 60 & 0 & 0\\
Qwen3.8-27B & Groq & 313 & 146 & 0 & 58 & 0\\
Gemma 4 26B-A4B$^\dagger$ & Gemini & 310 & 137 & 0 & 0 & 0\\
Sherkala-8B-Chat$^\dagger$ & open & 216 & 101 & 0 & 0 & 0\\
KazLLM-1.0-8B$^\dagger$ & open & 215 & 101 & 0 & 0 & 0\\
NLLB-200 (600M)$^\ddagger$ & open & 0 & 0 & 0 & 198 & 0\\
\midrule Human & -- & \multicolumn{5}{l}{3,811 passages (4 domains)}\\
\bottomrule\end{tabular}
\caption{Generators and passage counts after QC. M: machine text (C1, C2); MIX: human opening + machine continuation; HP: AI-polished human text; MT: machine translation from Russian (C4 LLM, C5 NLLB); MH: humanizer rewrite of M. Access: Gemini or Groq API, or open weights. $^\dagger$Held out: never used for training except in the last three rows of Table~\ref{tab:fixes}. Model cards: \citet{gemini31fl2026,gemini35fl2026,gemma4card2026,openai2025gptoss,qwen38card2026}. $^\ddagger$Gray-zone text only.}
\label{tab:gens}\end{table}

\paragraph{Stress tests.} On test-split passages we add: transliteration of both classes into the 2021 Latin-based alphabet (\textsc{Latin}); \res{const/hg-attack}\% of eligible Cyrillic letters in machine text replaced by Latin look-alikes (\textsc{Homoglyph}); human passages with naturally occurring homoglyphs, folded back to Cyrillic; and truncation of both classes to their first \res{const/trunc-long} and \res{const/trunc-short} words. Which passages carry each variant is a fixed function of the passage identifier. Development passages (\S\ref{sec:protocol}) are truncated the same way for length-specific calibration.

\paragraph{Normalization.} Both classes get the same normalization, and every detector reads only normalized text: Unicode NFC, markdown removal, one form each for quotes, dashes, spaces and ellipses, deletion of soft hyphens and zero-width characters, removal of a standalone first-line title, and a cut after the last sentence-final mark. Each step removes a cue that separates the classes for reasons unrelated to authorship (\S\ref{sec:findings}, Table~\ref{tab:cues}). Latin homoglyphs, which occur in real human Kazakh, are kept.

\paragraph{Validity audit.} \annotator{} judged \res{aud/n-human-judged} randomly ordered items (\res{x/aud1/n-h} human, \res{x/aud1/n-m} machine; claimed class shown): \res{aud/human-valid-h}\% of human and \res{aud/human-valid-m}\% of machine items were valid. The most common machine defect was text that reads like machine translation (\res{x/aud1/mt-m} of \res{x/aud1/n-m}; Appendix~\ref{app:audit}). Flagged human items and broken machine items were removed, as were \res{x/aud1/screen-only-h} human and \res{x/aud1/screen-only-m} machine passages flagged only by an LLM (Claude Opus 5.5) that read all \res{aud/n-llm} items drawn for audit and applied the same reasons, and the failure types became the corpus-wide filters above, which did not remove most such defects. After the benchmark was frozen, the same annotator judged a fresh random sample of \res{aud/post/n-h} human (\res{aud/post/n-h-per-domain} per domain) and \res{aud/post/n-m} machine passages (\res{aud/post/n-m-per-gen} per generator): only \res{aud/post/h-valid} human passages were valid (\res{aud/post/h-valid-w}\% weighted by domain size, \res{const/ci-level}\% CI [\res{aud/post/h-valid-w-ci/lo}, \res{aud/post/h-valid-w-ci/hi}]; educational \res{x/aud2/essay-valid} of \res{x/aud2/essay-n}), with \res{aud/post/h-broken} broken or truncated, \res{aud/post/h-mt} reading like machine translation and \res{aud/post/h-other} boilerplate or partly Russian; \res{aud/post/m-broken} machine passages were broken and \res{aud/post/m-mt} read like translation. We release the frozen benchmark with these judgments, so the human class is public pre-ChatGPT text as found. Public detectors do not flag only the defective passages: Fakespot flags \res{aud/post/fakespot-roberta/valid} and TMR \res{aud/post/tmr-roberta/valid} of the \res{aud/post/n-valid} valid ones.

\section{Detectors and Protocol}
\label{sec:protocol}
\paragraph{Zero-shot.} Each scoring family pairs a base model (observer) with a sibling that shares its tokenizer (performer): Qwen3-4B \citep{yang2025qwen3} and Llama-3.2-3B \citep{grattafiori2024llama3} with their instruction-tuned versions, Gemma-4-E2B with Gemma-4-E2B-it, and mGPT-1.3B \citep{shliazhko2024mgpt} with its Kazakh-adapted version \texttt{mGPT-1.3B-kazakh}. Gemma-4 is also a generator family; we flag that overlap. Table~\ref{tab:main} reports Binoculars, which uses both models, and Fast-DetectGPT's analytic criterion on the observer (for mGPT also on the Kazakh performer); Appendix~\ref{app:zs} reports all \res{x/zs/n-stats} statistics the scoring runs store. Scores use the first \res{const/max-tokens} tokens.
\paragraph{Fine-tuned.} XLM-R-base \citep{conneau2020xlmr} and mDeBERTa-v3-base \citep{he2023debertav3} are fine-tuned on train-split human vs.\ machine text (C1, C2) of the \res{corpus/n-train-gens} in-distribution generators for \res{const/epochs} epochs (lr \res{const/lr}, \res{const/max-tokens} tokens), over \res{const/seeds-digit} seeds whose probabilities are averaged (Appendix~\ref{app:extra}).
\paragraph{Off-the-shelf.} We run \res{ots/n} widely downloaded public AI-text detectors from the Hugging Face Hub as released, with their default \res{const/default-thr} threshold: \RESotsList{} \citep{solaiman2019release,guo2023hc3}.\footnote{Hugging Face model identifiers: \RESotsIDs{}.} All five are English-trained; this is what an institution without Kazakh-specific tools can download today. We also give them the most favourable setting, a threshold recalibrated on Kazakh development text.
\paragraph{Calibration and metrics.} We hold out \res{det/n/dev-clusters} of the \res{val/split/train-clusters} training topic clusters as a \emph{development} set that is never used for training: \res{det/n/H-dev} human passages (\res{val/split/dev-encyclopedic} encyclopedic, \res{val/split/dev-news} news, \res{val/split/dev-web} web; the random draw contains no educational cluster, so educational FPRs test a threshold set on other genres). Thresholds are set to \res{const/target-fpr}\% FPR on its human passages and applied unchanged to test, as a deployed tool would be configured before it meets new writers; for classifiers we also report the default \res{const/default-thr}. Zero-shot detectors never see training data and are calibrated on all training-split human passages. Length-specific thresholds, where used, apply to every input by its length, as in the released detector: the full-length threshold from 110 words, the 100-word one for 100--109 words and the 50-word one for 50--99. We report AUROC, FPR on test-split human text by domain and length, TPR at the operating threshold, and flag rates on gray classes, with \res{const/ci-level}\% CIs from a bootstrap over source documents (a human passage together with the machine and gray-zone texts prompted from it). For FPR at a calibrated threshold the bootstrap also resamples the development humans and re-fits the threshold, because with \res{det/n/H-dev} of them a nominal \res{const/target-fpr}\% threshold has a true FPR between \res{val/thr/lo}\% and \res{val/thr/hi}\% (\res{const/ci-level}\% of random samples; mean \res{val/thr/mean}\%). Held-out generators are scored on their outputs for test-split sources, so that configurations trained on them are compared on the same items. Gray-zone rates use MIX, HP and MH passages built from test-split sources only, since the others contain or rewrite training passages.

\section{Findings}
\label{sec:findings}
\begin{table*}[t]\centering\small\setlength{\tabcolsep}{4pt}
\resizebox{\textwidth}{!}{\begin{tabular}{lccc|ccccc|c|cc}\toprule
 & \multicolumn{2}{c}{AUROC} & & \multicolumn{6}{c|}{FPR on human test passages (\%)} & \multicolumn{2}{c}{TPR (\%)}\\
\cmidrule(lr){2-3}\cmidrule(lr){5-10}\cmidrule(lr){11-12}
Detector & in-dist.\ [95\% CI] & held-out & thr. & all & enc. & news & web & edu. & 50 words & in-dist. & held-out\\\midrule
\multicolumn{12}{l}{\textit{Zero-shot, threshold at 1\% FPR on training-split human text}}\\
Gemma-4-E2B / Binoculars & 0.931 [0.921, 0.940] & 0.981 & cal & 0.8 & 1.1 & 0.2 & 0.9 & 1.9 & 3.0 & 50.3 & 79.9\\
Gemma-4-E2B / Fast-DetectGPT & 0.761 [0.743, 0.777] & 0.817 & cal & 0.7 & 0.3 & 0.6 & 1.3 & 0.0 & 0.3 & 10.8 & 19.5\\
Llama-3.2-3B / Binoculars & 0.664 [0.646, 0.681] & 0.745 & cal & 1.4 & 2.6 & 0.4 & 0.6 & 3.2 & 4.3 & 3.6 & 9.3\\
Llama-3.2-3B / Fast-DetectGPT & 0.814 [0.799, 0.828] & 0.902 & cal & 1.0 & 2.1 & 0.6 & 0.6 & 0.5 & 0.8 & 16.5 & 29.9\\
Qwen3-4B / Binoculars & 0.509 [0.487, 0.527] & 0.558 & cal & 1.0 & 1.3 & 0.0 & 1.7 & 1.4 & 4.8 & 1.3 & 3.4\\
Qwen3-4B / Fast-DetectGPT & 0.475 [0.454, 0.493] & 0.525 & cal & 1.4 & 2.6 & 0.2 & 1.9 & 0.9 & 2.8 & 1.5 & 4.3\\
mGPT-1.3B / Fast-DetectGPT & 0.808 [0.794, 0.824] & 0.858 & cal & 1.0 & 1.1 & 0.9 & 1.7 & 0.0 & 1.0 & 10.0 & 26.7\\
mGPT-1.3B-kazakh / Fast-DetectGPT & 0.849 [0.835, 0.863] & 0.832 & cal & 0.6 & 0.8 & 0.2 & 0.4 & 1.4 & 3.6 & 9.8 & 18.1\\
\midrule
\multicolumn{12}{l}{\textit{Fine-tuned on AINA-KK train, default threshold}}\\
XLM-R-base (fine-tuned) & 0.998 [0.996, 0.999] & 0.982 & 0.5 & 9.9 & 5.0 & 11.1 & 14.2 & 6.9 & 21.1 & 99.9 & 97.3\\
mDeBERTa-v3-base (fine-tuned) & 0.998 [0.996, 0.999] & 0.984 & 0.5 & 7.6 & 6.1 & 7.9 & 10.3 & 4.2 & 24.4 & 100.0 & 94.8\\
\midrule
\multicolumn{12}{l}{\textit{Off-the-shelf public detectors, as released}}\\
Fakespot RoBERTa & 0.621 [0.604, 0.644] & 0.623 & 0.5 & 25.5 & 27.2 & 21.3 & 24.7 & 33.3 & 87.6 & 35.2 & 37.1\\
OpenAI RoBERTa (GPT-2 output) & 0.650 [0.630, 0.666] & 0.595 & 0.5 & 11.8 & 7.1 & 11.5 & 10.8 & 22.7 & 9.9 & 19.6 & 16.3\\
HC3 RoBERTa (ChatGPT) & 0.499 [0.476, 0.519] & 0.520 & 0.5 & 0.0 & 0.0 & 0.0 & 0.0 & 0.0 & 0.0 & 0.0 & 0.0\\
TMR RoBERTa & 0.292 [0.274, 0.311] & 0.390 & 0.5 & 81.4 & 80.2 & 88.7 & 80.0 & 70.4 & 14.2 & 52.2 & 63.8\\
Desklib DeBERTa & 0.537 [0.516, 0.556] & 0.518 & 0.5 & 59.3 & 48.3 & 56.8 & 61.5 & 79.6 & 82.0 & 68.6 & 65.6\\
\bottomrule\end{tabular}}
\caption{Detectors as an institution would deploy them. Human: 1,530 test-split passages (50 words: 394 of them cut to their first 50 words). In-distribution machine text: 1,274 passages from the four training generators; held-out: 442 passages from the four unseen generators on test-split sources. AUROC CIs: bootstrap over source documents.}\label{tab:main}\end{table*}

\paragraph{Public detectors accuse and miss.} Widely downloaded AI-text detectors on the Hugging Face Hub are what an institution without a Kazakh tool can deploy today, and on Kazakh they fail in both directions (Table~\ref{tab:main}, bottom). TMR RoBERTa flags \res{ots/tmr-roberta/fpr-at-0.5}\% of human passages and ranks human text above machine text (AUROC \res{ots/tmr-roberta/auroc}). Fakespot flags \res{ots/fakespot-roberta/fpr-at-0.5}\% of human passages, \res{ots/fakespot-roberta/fpr-by-domain-at-0.5/essay}\% of student reports and essays and \res{ots/fakespot-roberta/short/50/fpr-at-0.5}\% of \res{const/trunc-short}-word excerpts; Desklib DeBERTa flags \res{ots/desklib-deberta/fpr-by-domain-at-0.5/essay}\% of essays and the OpenAI detector \res{ots/openai-roberta/fpr-by-domain-at-0.5/essay}\%. If \res{const/prevalence}\% of submissions were machine-written (precision from the test FPR and the TPR pooled over all machine test passages), \res{fdr/ots-fakespot-roberta/0.5/5}\% of Fakespot's flags and \res{fdr/ots-tmr-roberta/0.5/5}\% of TMR's would fall on honest writers. Recalibration does not rescue them: at a threshold that gives \res{const/target-fpr}\% FPR on Kazakh development text, none catches more than \res{ots/max-tpr-dev}\% of machine passages, while four of them flag \res{ots/openai-roberta/fpr-at-dev}--\res{ots/desklib-deberta/fpr-at-dev}\% of test humans. At a \res{const/target-fpr}\% operating point these English-trained detectors carry no usable signal on Kazakh.

\paragraph{Zero-shot detectors leave machine text unchecked.} Calibrated on training-split human text, every zero-shot detector keeps its error rate on unseen humans (test FPR \res{zs/min-fpr-dev}--\res{zs/max-fpr-dev}\%), but much of the machine text passes. The best, Binoculars with Gemma-4-E2B (AUROC \res{det/zs:gemma4-e2b:binoculars/auroc}), catches \res{det/zs:gemma4-e2b:binoculars/tpr-at-dev}\% of in-distribution machine passages; no other of the \res{x/zs/n-stats} statistics we computed catches more than \res{x/zs/second-tpr}\% (\res{x/zs/second-name}, at \res{x/zs/second-fpr}\% test FPR; Appendix~\ref{app:zs}). The scoring model matters more than the statistic: with Qwen3-4B, Binoculars and Fast-DetectGPT are at chance (AUROC \res{det/zs:qwen3-4b:binoculars/auroc} and \res{det/zs:qwen3-4b:obs-fdg/auroc}). Gemma-4-E2B's held-out AUROC (\res{det/zs:gemma4-e2b:binoculars/auroc-heldout}) is likely inflated by family overlap, since \res{corpus/n-heldout-gemma} held-out generators are Gemma-4 models. Kazakh continued pretraining of mGPT, a controlled comparison with the same architecture and tokenizer, raises Fast-DetectGPT from AUROC \res{det/zs:mgpt-1.3b:obs-fdg/auroc} to \res{det/zs:mgpt-1.3b:perf-fdg/auroc}, short of the larger multilingual scorer.

\paragraph{Fine-tuned detectors rank well but mis-set the bar.} mDeBERTa fine-tuned on AINA-KK separates the classes (AUROC \res{det/base/auroc}; \res{det/base/auroc-heldout} on unseen generators), and not by length: on passages of \res{const/lenctrl-lo}--\res{const/lenctrl-hi} words only, AUROC is \res{det/base/auroc-len120-220}. Ranking is the easy part. A logistic regression on \res{val/surface/n-features} surface statistics, such as the spread of sentence lengths and the rates of punctuation and digits, reaches AUROC \res{val/surface/auroc} (\res{val/surface/auroc-len120-220} on \res{const/lenctrl-lo}--\res{const/lenctrl-hi}-word passages) and, calibrated the same way, catches \res{val/surface/tpr-at-dev}\% of in-distribution machine text at \res{val/surface/fpr-at-dev}\% test FPR, more than any zero-shot detector, though at about twice its FPR. Human passages are spans of published pages and machine passages complete compositions, so part of this separability is genre and form rather than authorship; ranking on AINA-KK is an upper bound for real submissions. Yet at the default \res{const/default-thr}, mDeBERTa flags \res{det/base/fpr-at-0.5}\% of human test passages and \res{det/base/fpr-by-domain-at-0.5/web}\% of web text, and XLM-R flags \res{det/xlmr/fpr-at-0.5}\%. The ranking transfers to unseen human writing; the threshold does not.

\begin{figure}[t]\centering\includegraphics[width=\columnwidth]{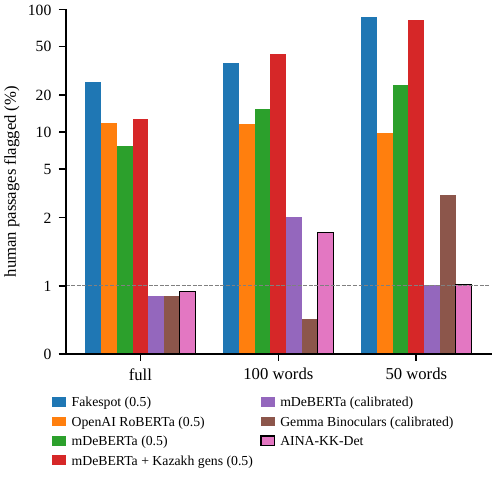}
\caption{Share of human test passages flagged at each detector's threshold, at full length and cut to their first \res{const/trunc-long} or \res{const/trunc-short} words (symmetric-log scale; dashed line: the \res{const/target-fpr}\% target). Both the default \res{const/default-thr} and a threshold calibrated on full-length text over-flag short text. AINA-KK-Det (\S\ref{sec:fixes}) uses length-matched training and length-specific thresholds.}\label{fig:length}\end{figure}

\paragraph{Short human texts are flagged most.} Cut to their first \res{const/trunc-short} words, human passages are flagged by mDeBERTa at \res{det/base/short/50/fpr-at-0.5}\% at the default threshold, against \res{det/base/fpr-at-0.5}\% at full length (Figure~\ref{fig:length}). A threshold calibrated on full-length text brings mDeBERTa to \res{det/base/short/50/fpr-at-dev}\%, but its script-augmented variant still flags \res{det/aug/short/50/fpr-at-dev}\%. The evidence points partly to the training data. Machine outputs are shorter than human passages (median \res{len/m-median} vs.\ \res{len/h-median} words; \res{len/m-under120}\% vs.\ \res{len/h-under120}\% under \res{const/span-min} words), so brevity becomes evidence of machine authorship. Sherkala is the shortest writer (\res{len/sherkala-under120}\% of outputs under \res{const/span-min} words), and adding KazLLM and Sherkala to the same recipe raises the \res{const/trunc-short}-word FPR at \res{const/default-thr} to \res{det/kznoaug/short/50/fpr-at-0.5}\%.

\begin{table*}[t]\centering\small\setlength{\tabcolsep}{4pt}
\resizebox{\textwidth}{!}{\begin{tabular}{l|cccc|cc|cc|cc}\toprule
 & \multicolumn{4}{c|}{Gray zones: flagged (\%)} & \multicolumn{2}{c|}{Kazakh-built: TPR} & \multicolumn{2}{c|}{Latin-2021} & \multicolumn{2}{c}{Homoglyphs: TPR} \\
Detector & HP & MT & MIX & MH & KazLLM & Sherkala & AUROC & TPR & before & after\\\midrule
Gemma-4-E2B / Binoculars & 7.7 & 8.2 & 35.3 & 20.3 & 50.6 & 74.7 & 0.346 & 0.0 & 49.1 & 12.5\\
mGPT-1.3B-kazakh / Fast-DetectGPT & 5.1 & 0.3 & 10.0 & 3.1 & 27.1 & 40.0 & 0.480 & 0.0 & 9.8 & 0.0\\
XLM-R-base (FT) & 6.4 & 17.0 & 68.8 & 98.4 & 69.4 & 64.0 & 0.875 & 0.0 & 99.3 & 27.1\\
mDeBERTa-v3-base (FT) & 7.7 & 7.9 & 82.5 & 100.0 & 47.1 & 62.7 & 0.384 & 0.0 & 99.0 & 63.3\\
\bottomrule\end{tabular}}
\caption{Stress tests at the calibrated threshold (1\% FPR on unseen human text). HP: AI-polished human text; MT: machine-translated from Russian; MIX: human opening ($\sim$50 words) continued by a model; MH: machine text rewritten by a second model to sound human. Latin-2021: both classes transliterated (503 human + 398 machine passages). Homoglyphs: 409 machine passages before and after 10\% of eligible Cyrillic letters are replaced by Latin look-alikes.}\label{tab:stress}\end{table*}

\paragraph{Kazakh-built models evade.} At the calibrated threshold, mDeBERTa catches at least \res{flag/base/min-global}\% of the passages of every global generator, including the \res{corpus/n-heldout-gemma} unseen Gemma-4 models, but only \res{det/base/flag-by-gen-at-dev/kazllm}\% of KazLLM (\res{x/ci/base/kazllm/k} of \res{x/ci/base/kazllm/n}; 95\% CI \res{x/ci/base/kazllm/cp/lo}--\res{x/ci/base/kazllm/cp/hi}) and \res{det/base/flag-by-gen-at-dev/sherkala}\% of Sherkala passages (\res{x/ci/base/sherkala/k} of \res{x/ci/base/sherkala/n}; \res{x/ci/base/sherkala/cp/lo}--\res{x/ci/base/sherkala/cp/hi}; Table~\ref{tab:stress}). Zero-shot Binoculars fails elsewhere: it catches \res{det/zs:gemma4-e2b:binoculars/flag-by-gen-at-dev/kazllm}\% of KazLLM and \res{det/zs:gemma4-e2b:binoculars/flag-by-gen-at-dev/sherkala}\% of Sherkala passages, and \res{det/complement/base/bino-catches}\% of the \res{det/complement/base/n-missed} Kazakh-built passages that mDeBERTa misses, so the two kinds of detector err on different texts (\S\ref{sec:fixes}). A classifier trained on four global generators misses much of the output of two Kazakh-adapted Llama-3.1-8B models, although their shorter outputs should make them easier to flag; size, base model and local decoding are confounded with being Kazakh-built.

\paragraph{Latin-2021 transliteration breaks every calibrated threshold.} Transliterated into the 2021 Latin alphabet, the ranking inverts for mDeBERTa (AUROC \res{det/base/script/latin2021}) and Gemma Binoculars (\res{det/zs:gemma4-e2b:binoculars/script/latin2021}) and survives only partly for XLM-R (\res{det/xlmr/script/latin2021}). At their calibrated thresholds, \res{val/latin/n-low} of the \res{val/latin/n} detectors we ran without script handling (Appendix~\ref{app:zs}) catch at most \res{val/latin/max-tpr-low}\% of Latin machine text, mDeBERTa included, and the other \res{val/latin/n-high} catch more only by also flagging \res{val/latin/min-fpr-high}--\res{val/latin/max-fpr-high}\% of Latin human text. Replacing \res{const/hg-attack}\% of eligible Cyrillic letters with Latin look-alikes cuts mDeBERTa's TPR from \res{det/base/script/hg-tpr-orig-at-dev}\% to \res{det/base/script/hg-tpr-at-dev}\%. Flagging homoglyphs is no remedy, because \res{x/hg/all}\% of human passages (\res{x/hg/test}\% in the test split) contain them.

\paragraph{Gray zones have no stable answer.} At the calibrated threshold mDeBERTa flags \res{det/base/flag-gray-at-dev/MIX}\% of passages that open with $\sim$\res{const/c2-prefix} human words and continue with machine text, \res{det/base/flag-gray-at-dev/HP}\% of AI-polished human passages and \res{det/base/flag-gray-at-dev/MT}\% of machine translations from Russian. Our prompt-based humanizer does not evade (\res{det/base/flag-gray-at-dev/MH}\% flagged), probably because its rewriting model is also a training generator. Whether such text should be flagged is a policy decision, and a flag says little about how much of a passage a person wrote.

\paragraph{Surface cues.} Before normalization, characters that almost no human passage contains mark machine text (Table~\ref{tab:cues}), which is why both classes are normalized identically. Content differences remain: \res{val/digits/h}\% of human test passages contain a digit, against \res{val/digits/m}\% of in-distribution machine passages. Among passages without digits mDeBERTa still ranks well (AUROC \res{val/digits/base/auroc-nodigit}), but its threshold, set mostly on digit-bearing text, flags \res{val/digits/base/fpr-nodigit}\% of the \res{val/digits/n-h-nodigit} digit-free human passages, \res{val/digits/n-h-nodigit-essay} of them educational, against \res{val/digits/base/fpr-digit}\% of the rest.

\section{Fixes and AINA-KK-Det}
\label{sec:fixes}
Each fix below uses only data that AINA-KK already contains. Table~\ref{tab:fixes} applies them cumulatively to mDeBERTa.

\begin{table*}[t]\centering\small\setlength{\tabcolsep}{4pt}
\resizebox{\textwidth}{!}{\begin{tabular}{l|ccc|ccc|cc|cc}\toprule
 & \multicolumn{3}{c|}{FPR on human text (\%)} & \multicolumn{3}{c|}{TPR (\%)} & \multicolumn{2}{c|}{Kazakh-built: TPR} & \multicolumn{2}{c}{Script: TPR} \\
Configuration & all [95\% CI] & web & 50 words & in-dist. & held-out$^\ast$ & 50 words & KazLLM & Sherkala & Latin & homoglyph\\\midrule
mDeBERTa, threshold 0.5 & 7.6 & 10.3 & 24.4 & 100.0 & 94.8 & 99.5 & 85.9 & 85.3 & -- & --\\
\quad + threshold calibrated on unseen humans & 0.8 [0.1, 3.7] & 1.7 & 1.0 & 98.8 & 83.5 & 67.5 & 47.1 & 62.7 & 0.0 & 63.3\\
\quad + script canonicalization & 0.8 [0.1, 3.7] & 1.7 & 1.0 & 98.8 & 83.5 & 67.5 & 47.1 & 62.7 & 99.0 & 99.0\\
\quad + script-augmented training & 0.8 [0.0, 2.9] & 1.7 & 5.6 & 99.7 & 84.6 & 95.7 & 48.2 & 68.0 & 99.7 & 99.8\\
\quad + length-matched training & 1.9 [0.4, 3.2] & 3.2 & 3.6 & 99.8 & 84.8 & 94.2 & 48.2 & 69.3 & 99.7 & 99.8\\
\quad + length-specific thresholds & 1.9 [0.4, 3.2] & 3.2 & 1.3 & 98.7 & 82.6 & 67.3 & 45.9 & 61.3 & 99.2 & 99.0\\
\quad + OR Gemma-4-E2B Binoculars (\textbf{AINA-KK-Det}) & 0.9 [0.4, 2.7] & 1.5 & 1.0 & 85.2 & 81.4 & 41.1 & 57.6 & 65.3 & 85.2 & 84.1\\
\midrule
mDeBERTa + KazLLM and Sherkala in training & 1.0 [0.3, 3.7] & 2.2 & 6.3 & 98.7 & 91.9 & 84.3 & 76.5 & 80.0 & 90.2 & 98.8\\
\quad + script-augmented training & 1.2 [0.1, 3.3] & 2.8 & 0.8 & 99.5 & 86.0 & 23.4 & 57.6 & 65.3 & 98.7 & 99.5\\
\quad + length-matched, length-specific thresholds & 0.8 [0.1, 2.0] & 2.2 & 1.8 & 64.8 & 49.3 & 54.1 & 41.2 & 44.0 & 72.4 & 63.3\\
\bottomrule\end{tabular}}
\caption{Fixes, applied cumulatively to mDeBERTa-v3-base (3 seeds, probabilities averaged). Calibrated thresholds give 1\% FPR on development humans (391 passages from training topics held out of training); length-specific thresholds are fitted on the same passages cut to 100 or 50 words and applied to every input by its length (\S\ref{sec:protocol}). Script canonicalization maps Latin-2021 text back to Cyrillic and folds Latin homoglyphs before scoring. OR: a text is flagged if either detector exceeds its own threshold, the two set jointly to 1\% FPR on development humans. $^\ast$Held-out: test-source outputs of the four held-out generators; in the last three rows KazLLM and Sherkala outputs for training-split sources are also used for training, so only the two Gemma-4 models remain unseen there. FPR CI: bootstrap over source documents that also resamples the development humans and re-fits the thresholds, so it includes their sampling error. Appendix~\ref{app:devrep} repeats this table on development data alone.}\label{tab:fixes}\end{table*}

\paragraph{Calibrate on unseen human text.} A threshold set to \res{const/target-fpr}\% FPR on development humans, from training topics withheld from training, gives \res{det/base/fpr-at-dev}\% FPR on test humans [\res{det/base/fpr-at-dev-ci-thr/lo}, \res{det/base/fpr-at-dev-ci-thr/hi}] and keeps TPR at \res{det/base/tpr-at-dev}\% for training generators and \res{det/base/tpr-heldout-at-dev}\% for unseen ones. Per \res{const/per-n} human passages, false flags fall from \res{falseflags/base/0.5/per1000} to \res{falseflags/base/dev/per1000} (95\% CI \res{x/falseflags/base/dev/lo}--\res{x/falseflags/base/dev/hi}). All this needs is a held-out sample of local human writing, and a claimed false-positive rate means little without one. Single fine-tuned models are unstable at this operating point (in-distribution TPR \res{seedspan/aug/tpr-at-dev}\% across seeds with script augmentation; Appendix~\ref{app:seeds}), so every fine-tuned result averages three seeds' probabilities.

\paragraph{Canonicalize the script before scoring.} We map Latin-2021 text back to Cyrillic with a rule-based inverse transliterator (a round trip of the \res{canon/n} transliterated test passages restores \res{canon/word-exact}\% of words exactly; this measures what the 2021 alphabet loses, not accuracy on Latin text written by people) and fold Latin homoglyphs into Cyrillic. Without retraining, mDeBERTa's TPR on Latin machine text rises from \res{det/base/script/latin-tpr-at-dev}\% to \res{det/base/script/canon-tpr-at-dev}\%, and on homoglyph-attacked text from \res{det/base/script/hg-tpr-at-dev}\% to \res{det/base/script/hgfold-tpr-at-dev}\%, while FPR on canonicalized human text is \res{det/base/script/canon-fpr-at-dev}\%. Gemma Binoculars' Latin AUROC rises from \res{det/zs:gemma4-e2b:binoculars/script/latin2021} to \res{det/zs:gemma4-e2b:binoculars/script/latin2021canon}.

\paragraph{Train with script augmentation.} Transliterating \res{const/aug-latin-words} of training passages to Latin-2021 and adding homoglyphs to \res{const/aug-hg-words} makes the model robust by itself (Latin AUROC \res{det/aug/script/latin2021}) and lowers FPR at \res{const/default-thr} from \res{det/base/fpr-at-0.5}\% to \res{det/aug/fpr-at-0.5}\%. Canonicalization still adds detection at the calibrated threshold (Latin TPR \res{det/aug/script/latin-tpr-at-dev}\% without it, \res{det/aug/script/canon-tpr-at-dev}\% with it), so we use both.

\paragraph{Match lengths.} In the spirit of multiscale training \citep{tian2024multiscale}, but aimed at the human class, we cut \res{const/lenmatch-h} of the human training passages to lengths drawn from the machine-output distribution, and \res{const/lenmatch-ex}\% of both classes to \res{const/excerpt-lo}--\res{const/excerpt-hi}-word excerpts. This weakens brevity as a cue: at \res{const/default-thr}, the \res{const/trunc-short}-word FPR falls from \res{det/aug/short/50/fpr-at-0.5}\% to \res{det/auglen/short/50/fpr-at-0.5}\%, and at the full-length calibrated threshold from \res{det/aug/short/50/fpr-at-dev}\% to \res{det/auglen/short/50/fpr-at-dev}\%. Thresholds fitted on development humans cut to the input length bring it to \res{det/auglen/short/50/fpr-at-devlen}\% [\res{det/auglen/short/50/fpr-at-devlen-ci-thr/lo}, \res{det/auglen/short/50/fpr-at-devlen-ci-thr/hi}] at \res{det/auglen/short/50/tpr-at-devlen}\% TPR on \res{const/trunc-short}-word machine excerpts; they do most of this work, since without length-matched training they give \res{det/aug/short/50/fpr-at-devlen}\% at \res{det/aug/short/50/tpr-at-devlen}\% TPR. Applied to every input by its length, they also lower recall on naturally short outputs: in-distribution TPR \res{x/dep/auglen/tpr}\% instead of \res{det/auglen/tpr-at-dev}\%, and KazLLM and Sherkala \res{x/dep/auglen/kazllm}\% and \res{x/dep/auglen/sherkala}\% instead of \res{det/auglen/flag-by-gen-at-dev/kazllm}\% and \res{det/auglen/flag-by-gen-at-dev/sherkala}\%. The full-length test FPR rises to \res{det/auglen/fpr-at-dev}\% [\res{det/auglen/fpr-at-dev-ci-thr/lo}, \res{det/auglen/fpr-at-dev-ci-thr/hi}], but on development data only from \res{devrep/aug/fpr}\% to \res{devrep/auglen/fpr}\% (Appendix~\ref{app:devrep}), so this cost may be threshold noise. \res{const/trunc-short-word} words remain harder than full passages, so a deployed tool should report the error rate for the input's length with each score.

\paragraph{Training on Kazakh-built generators is not a dependable fix.} Adding KazLLM and Sherkala outputs for training-split sources, without script augmentation, raises their detection at the calibrated threshold to \res{det/kznoaug/flag-by-gen-at-dev/kazllm}\% and \res{det/kznoaug/flag-by-gen-at-dev/sherkala}\% (from \res{det/base/flag-by-gen-at-dev/kazllm}\% and \res{det/base/flag-by-gen-at-dev/sherkala}\%). At the default threshold, however, the same model flags \res{det/kznoaug/short/50/fpr-at-0.5}\% of \res{const/trunc-short}-word human excerpts and \res{det/kznoaug/fpr-by-domain-at-0.5/essay}\% of essays. With script augmentation the gain shrinks to \res{det/kz/flag-by-gen-at-dev/kazllm}\% and \res{det/kz/flag-by-gen-at-dev/sherkala}\%. With length matching and length-specific thresholds as well, the calibrated operating point breaks down: in-distribution TPR falls to \res{x/dep/kzlen/tpr}\% (single seeds at the full-length threshold \res{seedspan/kzlen/tpr-at-dev}\%), and KazLLM and Sherkala detection to \res{x/dep/kzlen/kazllm}\% and \res{x/dep/kzlen/sherkala}\%. Their output is short and close to human Kazakh text, so where a classifier puts the boundary depends on the recipe (Table~\ref{tab:fixes}).

\paragraph{Combine a classifier with a zero-shot scorer.} The length-matched classifier (Table~\ref{tab:fixes}, fifth row) catches at least \res{flag/auglen/min-global}\% of the text of every global generator, and Binoculars, at its own \res{const/target-fpr}\% threshold, catches \res{det/complement/auglen/bino-catches}\% of the \res{det/complement/auglen/n-missed} Kazakh-built passages that this classifier misses (\S\ref{sec:findings}). We flag a text when either detector exceeds its own threshold, the two set jointly to \res{const/target-fpr}\% FPR on development humans (each score is mapped to its empirical distribution on those humans and the maximum is thresholded). Over the classifier with length-specific thresholds, detection rises from \res{x/dep/auglen/kazllm}\% to \res{x/dep/combo/kazllm}\% for KazLLM and from \res{x/dep/auglen/sherkala}\% to \res{x/dep/combo/sherkala}\% for Sherkala: on the \res{x/dep/combo/kz/n} Kazakh-built test passages the combination catches \res{x/mc/auglen/only-second} that the classifier misses and misses \res{x/mc/auglen/only-first} that it catches (exact McNemar $p$=\res{x/mc/auglen/p}), and on development data KazLLM detection rises from \res{x/devdep/auglen/tpr-kazllm/k} to \res{x/devdep/combo-auglen/tpr-kazllm/k} of 27 (Table~\ref{tab:devrep}). The gain is consistent but not established; with the other two classifiers it is smaller ($p$=\res{x/mc/aug/p} and \res{x/mc/base/p}). Its lower test FPR (\res{x/dep/combo/fpr}\% [\res{x/dep/combo/fpr-ci/lo}, \res{x/dep/combo/fpr-ci/hi}] against \res{det/auglen/fpr-at-dev}\%) does not recur on development data (\res{devrep/combo/fpr}\% for both), so we do not credit it to the combination. The price is in-distribution recall, \res{x/dep/combo/tpr}\% instead of \res{x/dep/auglen/tpr}\%: the classifier's share of the joint threshold is set by the few development humans it scores near 1, the saturation seen across seeds.

\paragraph{AINA-KK-Det.} We release this configuration: script canonicalization; mDeBERTa trained with script augmentation and length matching, OR-combined with Gemma-4-E2B Binoculars; and length-specific thresholds. It was not chosen blind: the training recipes were developed on earlier versions of AINA-KK with the same human test passages, and OR-combinations of Binoculars with \res{det/n-combos} classifier variants, two trained on all eight generators, were compared on test. Its development-data rates (Table~\ref{tab:devrep}), where no configuration was trained, are therefore the less biased estimate: \res{x/devdep/combo/fpr}\% of human passages and \res{x/devdep/combo/fpr50}\% of \res{const/trunc-short}-word excerpts flagged, \res{x/devdep/combo/tpr-indist}\% of in-distribution machine passages caught. On test it flags \res{x/dep/combo/fpr}\% of human passages and \res{det/combo-auglen+bino/short/50/fpr-at-devlen}\% of \res{const/trunc-short}-word excerpts [\res{det/combo-auglen+bino/short/50/fpr-at-devlen-ci-thr/lo}, \res{det/combo-auglen+bino/short/50/fpr-at-devlen-ci-thr/hi}], and catches \res{x/dep/combo/tpr}\% of machine passages from training generators, \res{x/dep/combo/gemma4}\% from the unseen Gemma-4 models, \res{x/dep/combo/kz}\% from the Kazakh-built ones (95\% CI \res{x/dep/combo/kz/cp/lo}--\res{x/dep/combo/kz/cp/hi}) and \res{x/dep/combo/latin-canon}\% of Latin-script machine text. The script-augmented combination without length matching flags fewer test humans (\res{x/dep/combo-aug/fpr}\%) and catches more in-distribution text (\res{x/dep/combo-aug/tpr}\%) but fewer Kazakh-built passages (\res{x/dep/combo-aug/kz}\%; McNemar $p$=\res{x/mc/augcombo/p}). At \res{const/prevalence}\% prevalence, \res{x/dep/combo/ppv5}\% of AINA-KK-Det's flags would be correct (\res{x/dep/combo/ppv5-at-hi}\% at the upper end of its FPR interval), against \res{ppv/ots-fakespot-roberta/0.5/5}\% of Fakespot's. It flags \res{x/dep/combo/gray/HP}\% of AI-polished human text and \res{val/digits/combo/flagged-nodigit} of the \res{val/digits/n-h-nodigit} digit-free human test passages (\res{val/digits/combo/fpr-nodigit}\%, against \res{val/digits/combo/fpr-digit}\%), half of its false flags. We release it retrained with the same recipe and seeds, with its calibration data, its own measured rates (\res{x/release/tpr}\% in-distribution; \res{x/release/kazllm}\% KazLLM and \res{x/release/sherkala}\% Sherkala: a rerun moves these by several points) and the advice that a flag can start a conversation but is not evidence on its own.

\paragraph{Recommendations.} Do not use English-trained public detectors on Kazakh; calibrate on held-out human writing of the same genre and length and publish the FPR with an interval that includes the threshold's sampling error; canonicalize script and fit thresholds by length; and treat a flag as grounds for a conversation, never as proof.

\section{Conclusion}
Machine-generated Kazakh can be detected, but not with the tools available today. Downloaded English-trained detectors accuse honest writers and miss machine text, calibrated zero-shot detectors miss much of it, and fine-tuned detectors flag short human texts and break on transliterated text. Most of these failures have fixable causes: thresholds never calibrated on local human writing, a length gap between human and machine training text, and script. Kazakh-built generators remain the hardest case; pairing the classifier with a zero-shot scorer recovers part of their text at a cost in recall, a gain our test set cannot yet establish. We release AINA-KK, AINA-KK-Det and the calibration data, so that a claimed false-positive rate for Kazakh can be checked against real human writing.

\section*{Limitations}
(i) Generators are limited to public API tiers and open weights. The flagship models that students most use (ChatGPT's default model, Gemini Flash/Pro) are represented by lite or open siblings, and OpenAI only by its open-weight model. (ii) Human text is public text that existed before \res{const/cutoff-date}: encyclopedic, news, web and educational pages. Educational pages are student reports and essays as published online, not essays collected from students, and some are compilations; supervised L1/L2 writing sessions and a native-speaker detection study require ethics approval and are left to future work. Topic clusters, not outlets or websites, are split, so test FPRs describe new topics, not necessarily new sources. Scrubbed emails, phone and ID numbers leave placeholder tokens that only human text contains. (iii) Commercial detectors, including the local vendor whose claims motivate this work, are not evaluated: they offer no free API, and responsible evaluation means notifying vendors before publishing results. Of public checkpoints we ran \res{ots/n} English-trained ones, not every popular one; two with more downloads than four of ours, one of them XLM-R-based, were not run, and download rankings change from month to month. (iv) The human validity audits cover \res{aud/n-human-judged} items before and \res{aud/post/n} after filtering, by a single annotator who saw each item's claimed class. About a third of the sampled human passages remain defective, and only \res{x/aud2/essay-valid} of \res{x/aud2/essay-n} educational ones were valid, so FPRs describe public text as found rather than clean student writing, educational FPRs are the least certain, and the passages that read like machine translation may not be human-written at all. (v) Scoring models may have seen our human sources (Wikipedia, CC100) during pretraining, which can make human text look machine-like to zero-shot detectors. (vi) The Latin-script test uses machine-transliterated Cyrillic text with an exact inverse, not text written in Latin by people; informal Latin spellings and the 2017 and 2018 alphabets are not covered. Our humanizer is a prompt to a training generator rather than a commercial tool. (vii) In-distribution machine text comes from \res{corpus/n-train-gens} generators, so per-generator TPR estimates are coarse, and only \res{x/dep/combo/kazllm/n} KazLLM and \res{x/dep/combo/sherkala/n} Sherkala passages are test-split outputs (Appendix~\ref{app:ci}). The two Kazakh-built generators are both 8B Llama-3.1 derivatives decoded locally with a repetition penalty, so being Kazakh-built is confounded with size, base model and decoding. (viii) Thresholds are fitted on \res{det/n/H-dev} development passages from \res{det/n/dev-clusters} topic clusters, none of them educational, so the \res{const/target-fpr}\% quantile rests on a few tail passages; our FPR intervals include this, though they resample passages rather than topic clusters and so understate topic-level variation; single fine-tuned models vary widely across seeds at this operating point, so all fine-tuned results are \res{const/seeds}-seed averages. Educational test results rest on \res{val/split/essay-test-clusters} topic clusters. Human passages without digits are flagged more often (\S\ref{sec:findings}); genres with few numbers may see higher FPRs. (ix) The zero-shot component of AINA-KK-Det uses Gemma-4-E2B, which shares a model family with \res{corpus/n-heldout-gemma} held-out generators, and Gemma itself is built from the technology behind our Gemini training generators; we report those two separately. (x) The fixes were developed on earlier versions of the benchmark with the same human test passages, and AINA-KK-Det was chosen with test results in view. The development-only replication (Appendix~\ref{app:devrep}) is our check, but it rests on \res{devrep/n-h} human and \res{devrep/n-m} machine passages.

\section*{Ethics Statement}
No detector evaluated here, and no detector we are aware of, should be used as sole evidence of academic misconduct in Kazakh. All human text is public and was captured before ChatGPT's release (\res{const/cutoff-date}). We release Wikipedia text under CC BY-SA 3.0, the license of its pre-2023 revisions; news, educational and web passages are released as source identifiers (URLs, FineWeb2 and CC100 records) with the scripts that build them from the public sources, except that mixed-authorship passages contain their human opening (about \res{const/c2-prefix} words) and AI-polished passages are model rewrites of their source. The dataset is gated, and authors of source pages can request removal. Humanizer prompts and their outputs are released under the same gated access. AINA-KK-Det is released with its calibration data and its own measured false-positive rates, so that users can see what a flag does and does not mean.

\bibliography{refs}

\appendix
\section{Additional Results}
\label{app:extra}
\subsection{Surface cues and style}
\begin{table*}[t]\centering\footnotesize\setlength{\tabcolsep}{6pt}
\begin{tabular}{lrrl}\toprule
Cue & Human & Machine & Highest generator\\\midrule
non-breaking hyphen (U+2011) & 0.0 & 3.8 & gpt-oss-120b (41.7)\\
narrow no-break space & 0.0 & 0.7 & gpt-oss-120b (7.8)\\
soft hyphen & 1.6 & 0.1 & gpt-oss-120b (0.3)\\
ellipsis character & 2.2 & 0.3 & gpt-oss-120b (1.3)\\
standalone title line & 0.9 & 2.4 & Gemma 4 26B-A4B (6.7)\\
markdown header & 0.0 & 0.1 & gpt-oss-120b (0.9)\\
bold markup & 0.0 & 3.9 & Gemma 4 26B-A4B (10.2)\\
trailing fragment & 3.6 & 1.6 & Sherkala-8B-Chat (14.6)\\
mixed-script word & 0.0 & 1.6 & Gemini 3.5 Flash-Lite (4.0)\\
Latin homoglyph & 5.1 & 1.7 & Gemini 3.5 Flash-Lite (4.0)\\
\bottomrule\end{tabular}
\caption{Share of passages (\%) containing each surface cue before normalization: 3,299 encyclopedic, news and web passages before the second cleaning pass vs.\ 3,483 raw C1/C2 outputs.}\label{tab:cues}\end{table*}

Real Kazakh text and LLM output differ in ways unrelated to who wrote the content. Table~\ref{tab:cues} counts such cues before normalization. The non-breaking hyphen appears in \res{cue/nb_hyphen_U+2011/gptoss}\% of gpt-oss outputs, standalone title lines in \res{cue/title_line/gemma26}\% of Gemma~4 26B-A4B and \res{cue/title_line/gemma31}\% of Gemma~4~31B outputs, and soft hyphens and ellipsis characters almost only in human text. Gemini~3.5 Flash-Lite inserts Arabic, Hebrew, Greek, Georgian or Hangul letters inside Kazakh words in \res{cue/mixed_script_token/g35fl}\% of its outputs; we drop those outputs and report the rate as a generator defect.

\paragraph{Latin homoglyphs in human text.} \res{x/hg/all}\% of the benchmark's human passages contain Latin letters inside Cyrillic words (most often Latin \textit{i} for Kazakh \kk{і}; \res{cue/latin_homoglyph/h}\% of the Table~\ref{tab:cues} passages), against \res{cue/latin_homoglyph/m}\% of machine outputs. We keep them. Among the \res{det/base/script/hghum-n} test-split human passages that contain them, mDeBERTa flags \res{det/base/script/hghum-raw-at-dev}\% as written and \res{det/base/script/hghum-fold-at-dev}\% after folding, so canonicalization does not create false flags on real text.

\paragraph{Style.} Machine Kazakh differs from human Kazakh on simple features (\res{ling/n-human} human vs.\ \res{ling/n-machine} machine passages; Mann--Whitney tests with Benjamini--Hochberg correction). Its sentence lengths vary less (SD \res{ling/2/m} vs.\ \res{ling/2/h} words for human text; Cliff's $\delta=\res{ling/2/d}$), its vocabulary is more diverse (MATTR \res{ling/0/m} vs.\ \res{ling/0/h}; $\delta=\res{ling/0/d}$), it uses a higher share of Kazakh-specific letters ($\delta=\res{ling/4/d}$) and fewer \kk{-ған/-ген} participles ($\delta=\res{ling/7/d}$).

\paragraph{Kazakh-built outputs are not copies.} On average \res{overlap/kazllm/mean}\% (KazLLM) and \res{overlap/sherkala/mean}\% (Sherkala) of an output's word \res{const/ngram}-grams occur in the human passage it was prompted from.

\subsection{Leave-one-generator-out}
\begin{table}[t]\centering\small\setlength{\tabcolsep}{4pt}
\begin{tabular}{lcccc}\toprule
 & \multicolumn{2}{c}{AUROC} & \multicolumn{2}{c}{TPR (\%)}\\ \cmidrule(lr){2-3}\cmidrule(lr){4-5}
Generator & out & in & out & in\\\midrule
Gemini 3.1 Flash-Lite & 0.998 & 0.996 & 99.8 & 100.0\\
Gemini 3.5 Flash-Lite & 1.000 & 0.997 & 99.6 & 99.8\\
Qwen3.8-27B & 0.998 & 0.997 & 100.0 & 100.0\\
gpt-oss-120b & 0.997 & 0.996 & 94.4 & 99.2\\
\bottomrule\end{tabular}
\caption{Leave-one-generator-out among the four training generators. mDeBERTa with script augmentation and length matching, trained on three of them (out) or all four (in), scored on that generator's outputs for test-split sources against test humans; TPR at the calibrated 1\%-FPR threshold. One seed for out, three for in; single seeds are unstable at a 1\% threshold (\S\ref{sec:fixes}).}\label{tab:logo}\end{table}

Leaving one of the \res{corpus/n-train-gens} training generators out of training still leaves its text caught at no less than \res{logo/min-global-out}\% at the calibrated threshold (Table~\ref{tab:logo}). Transfer among these generators is not the problem; the Kazakh-built generators are (\S\ref{sec:findings}). The left-out models are single seeds, which \S\ref{sec:fixes} shows to be unstable at a \res{const/target-fpr}\% threshold, so this comparison is indicative.

\subsection{Development-only replication}
\label{app:devrep}
\begin{table*}[t]\centering\small\setlength{\tabcolsep}{5pt}
\resizebox{\textwidth}{!}{\begin{tabular}{l|ccc|cccc}\toprule
 & \multicolumn{3}{c|}{FPR on development humans (\%)} & \multicolumn{4}{c}{TPR on development machine text (\%)}\\
Configuration & all & web & 50 words & in-dist. & Gemma-4 & KazLLM & Sherkala\\\midrule
mDeBERTa, threshold 0.5 & 9.5 & 11.7 & 21.7 & 99.3 & 100.0 & 77.8 & 78.6\\
\quad + threshold calibrated on unseen humans & 1.3 & 0.0 & 2.3 & 95.6 & 100.0 & 33.3 & 57.1\\
\quad + script-augmented training & 1.3 & 0.0 & 6.4 & 98.5 & 100.0 & 33.3 & 57.1\\
\quad + length-matched training & 1.5 & 0.9 & 3.3 & 98.5 & 100.0 & 33.3 & 57.1\\
\quad + length-specific thresholds & 1.5 & 0.9 & 1.3 & 97.1 & 98.3 & 29.6 & 57.1\\
\quad + OR Gemma-4-E2B Binoculars (\textbf{AINA-KK-Det}) & 1.5 & 0.9 & 2.3 & 89.0 & 96.6 & 63.0 & 64.3\\
\midrule
mDeBERTa + KazLLM and Sherkala in training & 1.0 & 0.0 & 6.1 & 97.8 & 96.6 & 63.0 & 60.7\\
\quad + script-augmented training & 1.5 & 1.8 & 4.3 & 97.1 & 98.3 & 37.0 & 46.4\\
\quad + length-matched, length-specific thresholds & 2.8 & 4.5 & 2.0 & 70.6 & 63.8 & 37.0 & 46.4\\
\bottomrule\end{tabular}}
\caption{Table~\ref{tab:fixes} repeated on development data alone: 391 human passages from the 7 development topic clusters and the 249 machine passages prompted from them, none of which any configuration was trained on (the last three rows included). Each passage is judged with thresholds, and for the OR-combination the empirical distributions, fitted on the other six clusters; length-specific thresholds apply to every input by its length. Development data have no script variants, so the canonicalization row is omitted. Machine passages per column: 136, 58, 27 and 28.}\label{tab:devrep}\end{table*}

The fixes were developed on earlier versions of AINA-KK whose human test passages are the ones used here, so Table~\ref{tab:fixes} could reflect choices made with test results in view. Table~\ref{tab:devrep} repeats it on the development clusters, which no configuration was trained on. Their human passages set the thresholds of the main results, so here each cluster is judged with thresholds fitted on the other \res{devrep/n-fit-clusters}, and the FPRs come out somewhat above \res{const/target-fpr}\%, as expected for a quantile fitted on a few hundred passages (\S\ref{sec:protocol}). Rows with length-specific thresholds apply them to every input by its length, as in Table~\ref{tab:fixes}. The effects claimed in \S\ref{sec:fixes} reappear: calibration, the rise of the \res{const/trunc-short}-word FPR with script augmentation and its fall with length matching and length-specific thresholds, the combination trading in-distribution recall for Kazakh-built detection, and the collapse of in-distribution recall when KazLLM and Sherkala are added with length matching. Three test differences do not: length matching barely raises the full-length FPR here, the combination does not lower it, and in the last row Kazakh-built detection does not fall below the row above.

\subsection{Validity checks}
All \res{val/templates/n} prompts behind benchmark passages match exactly one template, from the pool of their source's split. The classifiers' training pools contain no development passage, no test-split passage and no output of a held-out generator; the last three rows of Table~\ref{tab:fixes} add \res{val/pools/kz-rows} KazLLM and Sherkala outputs for training-split sources and no Gemma-4 output. Every encyclopedic passage carries a revision timestamp (latest \res{val/dates/encyclopedic/max}) and every news and educational passage a Common Crawl capture date (latest \res{val/dates/news/max} for both); CC100 passages carry no per-page date and rest on the corpus's crawl period. A date before ChatGPT does not exclude earlier machine translation, which the validity audit looks for. In the released files, \texttt{split} gives a passage's role (held-out and gray-zone passages are all marked test) and \texttt{src\_split} the topic split of its source; analyses that need topic-disjoint sets should use \texttt{src\_split}.

\subsection{Prompts and decoding}
Prompts are written in Kazakh. Each condition has disjoint template pools for train- and test-split sources, so a detector cannot key on a template. One test template per condition, with an English gloss:
\begin{itemize}\itemsep0pt\small
\item \textbf{C1, encyclopedic}: \kk{Сіз энциклопедия авторысыз. «\{title\}» деген мақаланың кіріспе бөлімін шамамен \{n\} сөзбен жазыңыз\ldots} (``You are an encyclopedia author. Write the introduction of the article `\{title\}' in about \{n\} words\ldots'')
\item \textbf{C1, educational}: \kk{«\{cue\}» тақырыбында эссе жазып беріңізші. Шамамен \{n\} сөз болсын\ldots} (``Please write an essay on `\{cue\}', about \{n\} words\ldots'')
\item \textbf{C2}: \kk{Мәтіннің басы берілген. Оны табиғи түрде шамамен \{n\} сөзбен аяқтаңыз\ldots} (``The beginning of a text is given. Finish it naturally in about \{n\} words\ldots'')
\item \textbf{C3}: \kk{Мәтінді редакциялаңыз: қателерді түзетіп, оқуға жеңіл етіңіз, бірақ мазмұнын сақтаңыз\ldots} (``Edit the text: fix errors and make it easier to read, but keep its content\ldots'')
\item \textbf{C4}: \kk{Төмендегі мәтінді қазақшаға сапалы етіп аударыңыз\ldots} (``Translate the text below into Kazakh well\ldots'')
\item \textbf{C6}: \kk{Мәтінді өз сөзіңізбен қайта баяндаңыз, ол жасанды интеллект жазғандай болып көрінбесін\ldots} (``Retell the text in your own words so that it does not look AI-written\ldots'')
\end{itemize}
\{n\} is drawn at random from \res{const/req-words} words, around the human median of \res{len/h-median} words. Every generator writes less than asked (median \res{lenratio/min}--\res{lenratio/max}\% of the requested length), Sherkala least (\S\ref{sec:findings}). The released generation code samples each prompt at temperature 0.7 or 1.0 with top-$p$ 0.95 (the humanizer at 0.9); KazLLM and Sherkala run locally in float16 with repetition penalty 1.05, and NLLB-200 uses beam search with 4 beams.

\subsection{Seed averaging}
\label{app:seeds} A single fine-tuned model is fragile at a \res{const/target-fpr}\% operating point. Its probabilities saturate: in-distribution machine passages and the few highest-scoring development humans share a narrow band just below 1, so the threshold can land inside it. Across \res{const/seeds} seeds of the same recipe, in-distribution TPR at the calibrated threshold spans \res{seedspan/base/tpr-at-dev}\% for mDeBERTa and \res{seedspan/aug/tpr-at-dev}\% with script augmentation. Averaging the \res{const/seeds} seeds' probabilities narrows this (\res{det/base/tpr-at-dev}\% and \res{det/aug/tpr-at-dev}\%) but does not remove run-to-run variation (Table~\ref{tab:ci}); every fine-tuned result in this paper is such an average.

\subsection{Validity audit protocol}
\label{app:audit}
The annotator, a native speaker of Kazakh, saw each item's text and the class it was claimed to belong to (human or machine) and answered whether it is a valid example of that class. A negative answer took one of six reasons: broken text, website boilerplate, reads like machine translation, not Kazakh, model wrapper or refusal, or other. Of the \res{x/aud1/n-m} machine items, from \res{x/aud1/n-gens} generators, \res{x/aud1/mt-m} read like machine translation to the annotator (Gemma 4 31B \res{aud/mt/gemma31/n} of \res{aud/mt/gemma31/of}, gpt-oss-120b \res{aud/mt/gptoss/n} of \res{aud/mt/gptoss/of}, Gemini 3.1 Flash-Lite \res{aud/mt/g31fl/n} of \res{aud/mt/g31fl/of}, Qwen3.8-27B \res{aud/mt/qwen/n} of \res{aud/mt/qwen/of}). Human items judged invalid were removed, as were machine items judged broken or meta-text and the passages flagged by the LLM screen (\S\ref{sec:data}); translation-like machine outputs were kept, because that is what the generator produced.

\IfFileExists{results_screen.tex}{\makeatletter
\@namedef{res@scr/n}{100}
\@namedef{res@scr/batch}{25}
\@namedef{res@scr/n-batches}{4}
\@namedef{res@scr/model}{Claude Opus 5.5}
\@namedef{res@scr/n-h}{52}
\@namedef{res@scr/valid-h}{78.8}
\@namedef{res@scr/n-m}{48}
\@namedef{res@scr/valid-m}{89.6}
\@namedef{res@scr/n-both}{100}
\@namedef{res@scr/agree}{79.0}
\@namedef{res@scr/kappa}{0.41}
\@namedef{res@scr/date}{11 Oct 2026}
\makeatother
\paragraph{Fixed-prompt re-run of the LLM screen.} The screen used during construction had no fixed prompt, so after the benchmark was frozen we re-ran it with the annotator's instructions on the \res{scr/n} items the annotator judged, in the same order and with the same claimed class shown: \res{scr/model} in the claude.ai chat interface (incognito, default settings, \res{scr/date}), in \res{scr/n-batches} batches of \res{scr/batch} items, each in a new conversation, with one JSON line returned per item (prompt below; \texttt{scripts/llm\_screen.py} builds the batches and scores the replies). It judged \res{scr/valid-h}\% of human and \res{scr/valid-m}\% of machine items valid and agrees with the annotator on \res{scr/agree}\% of items (Cohen's $\kappa$=\res{scr/kappa}). The chat interface does not fix the sampling temperature, so a repeated run can differ slightly. The re-run does not change the benchmark: passages removed during construction stay removed, and no other passage was removed.
\begin{quote}\small\raggedright
You are annotating Kazakh texts for a research dataset. You will receive 25 items. Each item has an ID, a label and a text. The label says what the text is: HUMAN (written by people and published online before 30 November 2022), MACHINE (a language model's answer to a prompt), AI-POLISHED HUMAN (a human text edited by a language model) or MACHINE-TRANSLATED (translated into Kazakh by a machine). The label is correct; do not check it.\\
For each item, answer one question. For HUMAN items: is this a normal Kazakh text that can be kept as it is? For all other items: is this a normal model answer in Kazakh, without junk?\\
Answer yes unless one of these problems applies; then answer no and give that reason:\\
broken: cut off mid-word, letters of Hebrew, Arabic, Greek or other scripts inside words, leftover markup, or garbled text.\\
boilerplate: website junk such as menus, ``Home »'', lists of links, ``Download'', or a reference list.\\
looks\_mt: looks like machine translation, with calques from Russian or incoherent phrases.\\
not\_kazakh: mostly Russian or another language.\\
meta\_refusal: model wrapper, such as a title line at the start, ``\kk{Міне, мақала:}'', ``Here is\ldots{}'', or a refusal.\\
other: a problem not in this list.\\
These are still yes: typos and odd style, factual errors, Latin i, ə or ğ inside Kazakh words, and text that sounds like AI. If unsure, answer yes.\\
Judge each item on its own. Output one line per item, in the given order, in exactly this format and nothing else:\\
\texttt{\{"id": "I1", "valid": true, "reason": null\}}\\
\texttt{\{"id": "I2", "valid": false, "reason": "broken"\}}\\
\texttt{\#\#\# I1 | HUMAN}, then the text; one such block per item.
\end{quote}

}{}

\subsection{Training and compute}
Fine-tuned detectors use \res{const/epochs} epochs, AdamW (learning rate \res{const/lr}, weight decay \res{const/wd}, \res{const/warmup}\% linear warm-up), batch \res{const/batch} (\res{const/microbatch} with \res{const/accum}-step accumulation), \res{const/max-tokens} tokens, mixed precision with fp32 master weights, and gradient clipping at \res{const/clip}; \res{const/seeds} seeds per configuration, probabilities averaged. Script augmentation replaces a training passage with its Latin-2021 transliteration with probability \res{const/aug-latin} and adds Latin homoglyphs at a rate of \res{const/aug-hg-rates}\% with probability \res{const/aug-hg}. Zero-shot scores use the first \res{const/max-tokens} tokens. The fine-tuned encoders have about \res{ext/encoder-params}B parameters each and the zero-shot scorers \res{ext/zs-params} parameters. Fine-tuning used one \res{const/gpu-gb}\,GB GPU, and zero-shot scoring placed the observer and performer models on two such GPUs. One fine-tuning run takes about \res{compute/ft-run-minutes} minutes including scoring, and all experiments, including generation with the Kazakh-built models, took about \res{compute/gpu-hours} hours of GPU sessions.

\subsection{All zero-shot statistics}
\label{app:zs}
\begin{table*}[t]\centering\small\setlength{\tabcolsep}{5pt}
\begin{tabular}{llcccc}\toprule
 & Statistic & Gemma-4-E2B & Llama-3.2-3B & mGPT-1.3B & Qwen3-4B\\\midrule
 & Binoculars & \res{x/zs/gemma4-e2b:binoculars/tpr} (\res{x/zs/gemma4-e2b:binoculars/auroc}) & \res{x/zs/llama3.2-3b:binoculars/tpr} (\res{x/zs/llama3.2-3b:binoculars/auroc}) & \res{x/zs/mgpt-1.3b:binoculars/tpr} (\res{x/zs/mgpt-1.3b:binoculars/auroc}) & \res{x/zs/qwen3-4b:binoculars/tpr} (\res{x/zs/qwen3-4b:binoculars/auroc})\\
\multicolumn{6}{l}{\textit{observer}}\\
 & log-likelihood & \res{x/zs/gemma4-e2b:obs_ll/tpr} (\res{x/zs/gemma4-e2b:obs_ll/auroc}) & \res{x/zs/llama3.2-3b:obs_ll/tpr} (\res{x/zs/llama3.2-3b:obs_ll/auroc}) & \res{x/zs/mgpt-1.3b:obs_ll/tpr} (\res{x/zs/mgpt-1.3b:obs_ll/auroc}) & \res{x/zs/qwen3-4b:obs_ll/tpr} (\res{x/zs/qwen3-4b:obs_ll/auroc})\\
 & log-rank & \res{x/zs/gemma4-e2b:obs_logrank/tpr} (\res{x/zs/gemma4-e2b:obs_logrank/auroc}) & \res{x/zs/llama3.2-3b:obs_logrank/tpr} (\res{x/zs/llama3.2-3b:obs_logrank/auroc}) & \res{x/zs/mgpt-1.3b:obs_logrank/tpr} (\res{x/zs/mgpt-1.3b:obs_logrank/auroc}) & \res{x/zs/qwen3-4b:obs_logrank/tpr} (\res{x/zs/qwen3-4b:obs_logrank/auroc})\\
 & entropy & \res{x/zs/gemma4-e2b:obs_entropy/tpr} (\res{x/zs/gemma4-e2b:obs_entropy/auroc}) & \res{x/zs/llama3.2-3b:obs_entropy/tpr} (\res{x/zs/llama3.2-3b:obs_entropy/auroc}) & \res{x/zs/mgpt-1.3b:obs_entropy/tpr} (\res{x/zs/mgpt-1.3b:obs_entropy/auroc}) & \res{x/zs/qwen3-4b:obs_entropy/tpr} (\res{x/zs/qwen3-4b:obs_entropy/auroc})\\
 & likelihood/log-rank & \res{x/zs/gemma4-e2b:obs_lrr/tpr} (\res{x/zs/gemma4-e2b:obs_lrr/auroc}) & \res{x/zs/llama3.2-3b:obs_lrr/tpr} (\res{x/zs/llama3.2-3b:obs_lrr/auroc}) & \res{x/zs/mgpt-1.3b:obs_lrr/tpr} (\res{x/zs/mgpt-1.3b:obs_lrr/auroc}) & \res{x/zs/qwen3-4b:obs_lrr/tpr} (\res{x/zs/qwen3-4b:obs_lrr/auroc})\\
 & Fast-DetectGPT & \res{x/zs/gemma4-e2b:obs_fdg/tpr} (\res{x/zs/gemma4-e2b:obs_fdg/auroc}) & \res{x/zs/llama3.2-3b:obs_fdg/tpr} (\res{x/zs/llama3.2-3b:obs_fdg/auroc}) & \res{x/zs/mgpt-1.3b:obs_fdg/tpr} (\res{x/zs/mgpt-1.3b:obs_fdg/auroc}) & \res{x/zs/qwen3-4b:obs_fdg/tpr} (\res{x/zs/qwen3-4b:obs_fdg/auroc})\\
\multicolumn{6}{l}{\textit{performer}}\\
 & log-likelihood & \res{x/zs/gemma4-e2b:perf_ll/tpr} (\res{x/zs/gemma4-e2b:perf_ll/auroc}) & \res{x/zs/llama3.2-3b:perf_ll/tpr} (\res{x/zs/llama3.2-3b:perf_ll/auroc}) & \res{x/zs/mgpt-1.3b:perf_ll/tpr} (\res{x/zs/mgpt-1.3b:perf_ll/auroc}) & \res{x/zs/qwen3-4b:perf_ll/tpr} (\res{x/zs/qwen3-4b:perf_ll/auroc})\\
 & log-rank & \res{x/zs/gemma4-e2b:perf_logrank/tpr} (\res{x/zs/gemma4-e2b:perf_logrank/auroc}) & \res{x/zs/llama3.2-3b:perf_logrank/tpr} (\res{x/zs/llama3.2-3b:perf_logrank/auroc}) & \res{x/zs/mgpt-1.3b:perf_logrank/tpr} (\res{x/zs/mgpt-1.3b:perf_logrank/auroc}) & \res{x/zs/qwen3-4b:perf_logrank/tpr} (\res{x/zs/qwen3-4b:perf_logrank/auroc})\\
 & entropy & \res{x/zs/gemma4-e2b:perf_entropy/tpr} (\res{x/zs/gemma4-e2b:perf_entropy/auroc}) & \res{x/zs/llama3.2-3b:perf_entropy/tpr} (\res{x/zs/llama3.2-3b:perf_entropy/auroc}) & \res{x/zs/mgpt-1.3b:perf_entropy/tpr} (\res{x/zs/mgpt-1.3b:perf_entropy/auroc}) & \res{x/zs/qwen3-4b:perf_entropy/tpr} (\res{x/zs/qwen3-4b:perf_entropy/auroc})\\
 & likelihood/log-rank & \res{x/zs/gemma4-e2b:perf_lrr/tpr} (\res{x/zs/gemma4-e2b:perf_lrr/auroc}) & \res{x/zs/llama3.2-3b:perf_lrr/tpr} (\res{x/zs/llama3.2-3b:perf_lrr/auroc}) & \res{x/zs/mgpt-1.3b:perf_lrr/tpr} (\res{x/zs/mgpt-1.3b:perf_lrr/auroc}) & \res{x/zs/qwen3-4b:perf_lrr/tpr} (\res{x/zs/qwen3-4b:perf_lrr/auroc})\\
 & Fast-DetectGPT & \res{x/zs/gemma4-e2b:perf_fdg/tpr} (\res{x/zs/gemma4-e2b:perf_fdg/auroc}) & \res{x/zs/llama3.2-3b:perf_fdg/tpr} (\res{x/zs/llama3.2-3b:perf_fdg/auroc}) & \res{x/zs/mgpt-1.3b:perf_fdg/tpr} (\res{x/zs/mgpt-1.3b:perf_fdg/auroc}) & \res{x/zs/qwen3-4b:perf_fdg/tpr} (\res{x/zs/qwen3-4b:perf_fdg/auroc})\\
\bottomrule\end{tabular}
\caption{All \res{x/zs/n-stats} zero-shot statistics: TPR (\%) on in-distribution test machine text at a threshold giving \res{const/target-fpr}\% FPR on training-split humans, and AUROC in parentheses. Each statistic's direction is fixed on training-split data. Test FPRs range from \res{x/zs/min-fpr}\% to \res{x/zs/max-fpr}\%. For mGPT the performer is \texttt{mGPT-1.3B-kazakh}.}\label{tab:zsall}\end{table*}

The scoring runs store, for each family, log-likelihood, log-rank, entropy, their ratio \citep{su2023detectllm} and Fast-DetectGPT under both the observer and the performer, plus Binoculars. Table~\ref{tab:zsall} reports all of them, each oriented on training-split data and calibrated like Table~\ref{tab:main}. The \res{val/latin/n} detectors of the Latin-script count in \S\ref{sec:findings} are mDeBERTa with and without KazLLM and Sherkala in training, XLM-R, the five public detectors, and Binoculars and observer and performer Fast-DetectGPT for each family. Gemma-4-E2B Binoculars remains the best in-distribution detector; the next best, \res{x/zs/second-name}, catches \res{x/zs/second-tpr}\% at \res{x/zs/second-fpr}\% test FPR.

\subsection{Intervals for Kazakh-built detection}
\label{app:ci}
\begin{table*}[t]\centering\small\setlength{\tabcolsep}{4pt}
\resizebox{\textwidth}{!}{\begin{tabular}{l|ccc|ccc|ccc}\toprule
 & \multicolumn{3}{c|}{KazLLM} & \multicolumn{3}{c|}{Sherkala} & \multicolumn{3}{c}{Kazakh-built, pooled}\\
Detector & caught & exact CI & re-fit CI & caught & exact CI & re-fit CI & caught & exact CI & re-fit CI\\\midrule
mDeBERTa, calibrated & \res{x/ci/base/kazllm/k}/\res{x/ci/base/kazllm/n} & [\res{x/ci/base/kazllm/cp/lo}, \res{x/ci/base/kazllm/cp/hi}] & [\res{x/ci/base/kazllm/boot/lo}, \res{x/ci/base/kazllm/boot/hi}] & \res{x/ci/base/sherkala/k}/\res{x/ci/base/sherkala/n} & [\res{x/ci/base/sherkala/cp/lo}, \res{x/ci/base/sherkala/cp/hi}] & [\res{x/ci/base/sherkala/boot/lo}, \res{x/ci/base/sherkala/boot/hi}] & \res{x/ci/base/kz/k}/\res{x/ci/base/kz/n} & [\res{x/ci/base/kz/cp/lo}, \res{x/ci/base/kz/cp/hi}] & [\res{x/ci/base/kz/boot/lo}, \res{x/ci/base/kz/boot/hi}]\\
Gemma-4-E2B Binoculars & \res{x/ci/bino/kazllm/k}/\res{x/ci/bino/kazllm/n} & [\res{x/ci/bino/kazllm/cp/lo}, \res{x/ci/bino/kazllm/cp/hi}] & -- & \res{x/ci/bino/sherkala/k}/\res{x/ci/bino/sherkala/n} & [\res{x/ci/bino/sherkala/cp/lo}, \res{x/ci/bino/sherkala/cp/hi}] & -- & \res{x/ci/bino/kz/k}/\res{x/ci/bino/kz/n} & [\res{x/ci/bino/kz/cp/lo}, \res{x/ci/bino/kz/cp/hi}] & --\\
+ script aug., length matching and thresholds & \res{x/dep/auglen/kazllm/k}/\res{x/dep/auglen/kazllm/n} & [\res{x/dep/auglen/kazllm/cp/lo}, \res{x/dep/auglen/kazllm/cp/hi}] & [\res{x/dep/auglen/kazllm/boot/lo}, \res{x/dep/auglen/kazllm/boot/hi}] & \res{x/dep/auglen/sherkala/k}/\res{x/dep/auglen/sherkala/n} & [\res{x/dep/auglen/sherkala/cp/lo}, \res{x/dep/auglen/sherkala/cp/hi}] & [\res{x/dep/auglen/sherkala/boot/lo}, \res{x/dep/auglen/sherkala/boot/hi}] & \res{x/dep/auglen/kz/k}/\res{x/dep/auglen/kz/n} & [\res{x/dep/auglen/kz/cp/lo}, \res{x/dep/auglen/kz/cp/hi}] & [\res{x/dep/auglen/kz/boot/lo}, \res{x/dep/auglen/kz/boot/hi}]\\
AINA-KK-Det & \res{x/dep/combo/kazllm/k}/\res{x/dep/combo/kazllm/n} & [\res{x/dep/combo/kazllm/cp/lo}, \res{x/dep/combo/kazllm/cp/hi}] & [\res{x/dep/combo/kazllm/boot/lo}, \res{x/dep/combo/kazllm/boot/hi}] & \res{x/dep/combo/sherkala/k}/\res{x/dep/combo/sherkala/n} & [\res{x/dep/combo/sherkala/cp/lo}, \res{x/dep/combo/sherkala/cp/hi}] & [\res{x/dep/combo/sherkala/boot/lo}, \res{x/dep/combo/sherkala/boot/hi}] & \res{x/dep/combo/kz/k}/\res{x/dep/combo/kz/n} & [\res{x/dep/combo/kz/cp/lo}, \res{x/dep/combo/kz/cp/hi}] & [\res{x/dep/combo/kz/boot/lo}, \res{x/dep/combo/kz/boot/hi}]\\
\bottomrule\end{tabular}}
\caption{Detection of the Kazakh-built test outputs with 95\% intervals: exact (Clopper--Pearson, passages only) and a bootstrap over source documents that also resamples the development humans and re-fits the thresholds (as for FPRs). Rows 3--4 use length-specific thresholds on every input. The re-fit intervals of the single classifiers are wide because their thresholds sit in a band of saturated probabilities. Binoculars has no re-fit interval because its threshold is fitted on training-split humans, which this bootstrap does not resample.}\label{tab:ci}\end{table*}

Table~\ref{tab:ci} gives intervals for the smallest subsets the paper interprets: \res{x/dep/combo/kazllm/n} KazLLM and \res{x/dep/combo/sherkala/n} Sherkala test outputs. With passage sampling alone, the gap between global generators (at least \res{flag/base/min-global}\% caught) and Kazakh-built ones is far outside the intervals. With threshold re-fitting, the single classifiers' intervals span tens of points, while the combination's stay narrower.

\end{document}